\documentclass[10pt,twocolumn,letterpaper]{article}

\usepackage{wacv}
\usepackage{times}
\usepackage{epsfig}
\usepackage{graphicx}
\usepackage{amsmath}
\usepackage{amssymb}
\usepackage{xcolor}
\usepackage{comment}
\usepackage{enumitem}
\usepackage{multirow}
\usepackage{algorithm,algpseudocode}
\usepackage{wrapfig,booktabs}
\usepackage{diagbox,makecell}
\usepackage{bbm}
\usepackage{authblk}
\usepackage[accsupp]{axessibility} 
\newcommand{\myfirstpara}[1]{\noindent {\bf #1:}}
\newcommand{\mypara}[1]{\vspace{0.5em} \myfirstpara{#1}}

\def\wacvPaperID{1109} 
\wacvfinalcopy 

\ifwacvfinal
\fi

\ifwacvfinal
\usepackage[breaklinks=true,bookmarks=false]{hyperref}
\else
\usepackage[pagebackref=true,breaklinks=true,colorlinks,bookmarks=false]{hyperref}
\fi

\usepackage[capitalize]{cleveref}
\hypersetup{
    colorlinks=true,
    linkcolor=red,
    filecolor=magenta,      
    urlcolor=red,
    pdftitle={Overleaf Example},
    pdfpagemode=FullScreen,
    }
\begin{document}
\title{Does Data Repair Lead to Fair Models? \\ Curating Contextually Fair Data To Reduce Model Bias}

\author[1]{Sharat Agarwal \thanks{Equal Contribution}}
\author[2]{Sumanyu Muku$^\ast$}
\author[1]{Saket Anand}
\author[2]{Chetan Arora}

\affil[1]{IIIT Delhi, India {\tt\small \{sharata,anands\}@iiitd.ac.in}}
\affil[2]{Indian Institute of Technology Delhi, India {\tt\small \{muku95.cstaff@,chetan@cse.\}iitd.ac.in}}


\maketitle

\ifwacvfinal
\thispagestyle{empty}
\fi



\begin{abstract}
Contextual information is a valuable cue for Deep Neural Networks (DNNs) to learn better representations and improve accuracy. However, co-occurrence bias in the training dataset may hamper a DNN model's generalizability to unseen scenarios in the real world. For example, in COCO \cite{coco}, many object categories have a much higher co-occurrence with men compared to women, which can bias a DNN's prediction in favor of men. Recent works have focused on task-specific training strategies to handle bias in such scenarios, but fixing the available data is often ignored. In this paper, we propose a novel and more generic solution to address the contextual bias in the datasets by selecting a subset of the samples, which is fair in terms of the co-occurrence with various classes for a protected attribute. We introduce a data repair algorithm using the coefficient of variation($c_v$), which can curate fair and contextually balanced data for a protected class(es). This helps in training a fair model irrespective of the task, architecture or training methodology. Our proposed solution is simple, effective and can even be used in an active learning setting where the data labels are not present or being generated incrementally. We demonstrate the effectiveness of our algorithm for the task of object detection and multi-label image classification across different datasets. Through a series of experiments, we validate that curating contextually fair data helps make model predictions fair by balancing the true positive rate for the protected class across groups without compromising on the model's overall performance. Code:
\url{https://github.com/sumanyumuku98/contextual-bias}
\end{abstract}
\section{Introduction}

\begin{figure}[t]
\begin{center}
  \includegraphics[height=6.5cm, width = 0.8\linewidth]{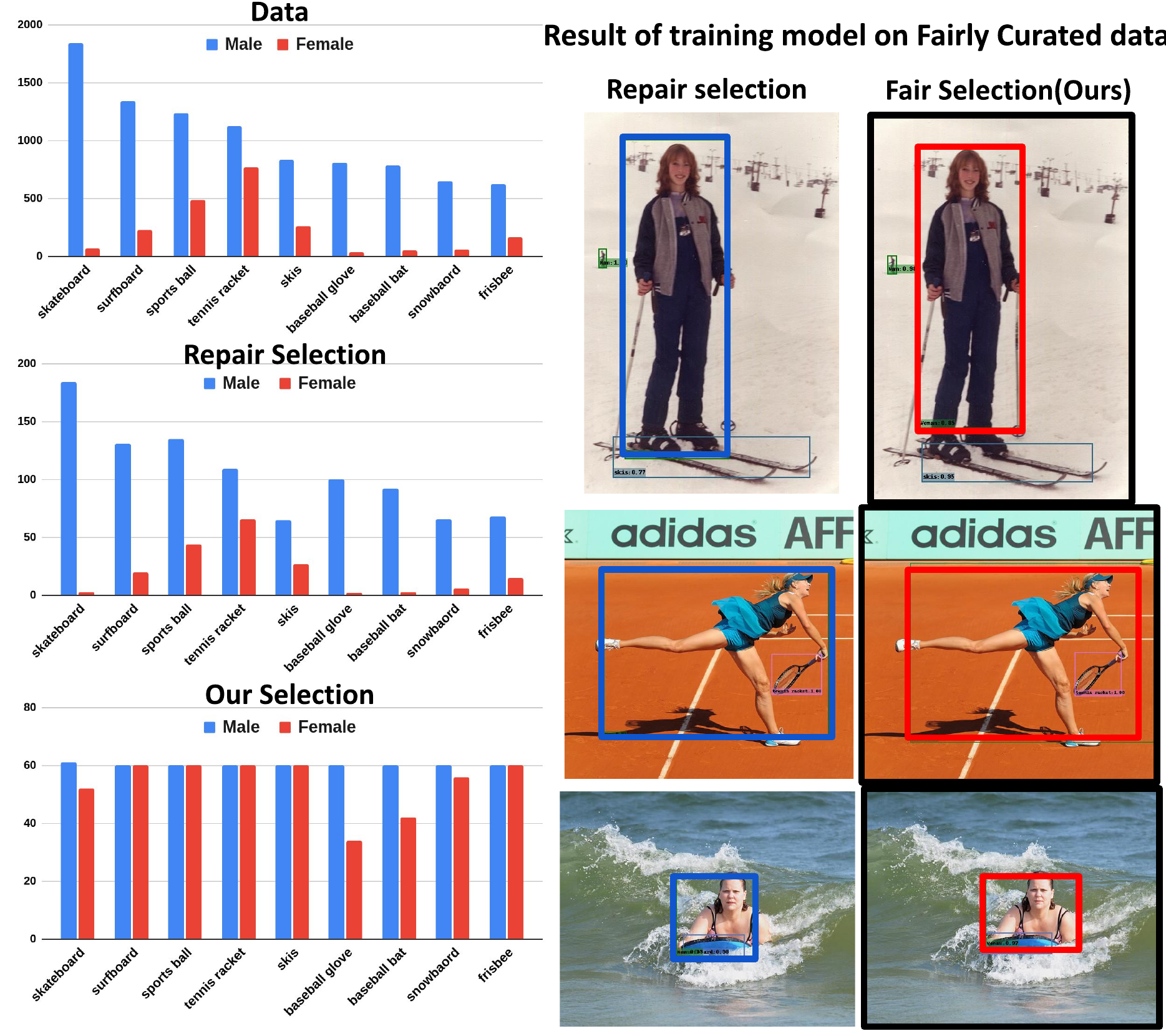}
  
\end{center}
   \caption{Bar plots in the left shows the count of men (blue) and women (red) for different sports objects in COCO dataset. We propose to curate a fair data for a protected class (gender in this case) across its co-occurring classes. As a result, the women which were detected as men with skis, tennis and surfboard in the middle column are now detected as women with high confidence in the right column, with only 20\% of the training data. The left-middle and left-bottom plots show the co-occurrence frequency after selecting a subset using Repair\cite{li2019repair}, and our technique respectively.}
\label{fig:teaser}
\end{figure}

Advancement in deep learning has mostly been model-centric, ignoring what it is working with, i.e., the data. A recent study \cite{paleyes2020challenges} reports that \emph{data} is the most overlooked and critical aspect of deep learning. By retraining the models with well curated data can improve the quality of learning in a much shorter time. Notably, large datasets are integral to enhancing the generalization performance of deep learning models \cite{2011_efors}. However, the challenge lies in annotating larger datasets, making them difficult to employ in the real world. Moreover, if the data is inadvertently skewed, the models will typically amplify the ill effects. Torralba and Efros \cite{2011_efors} pointed out that most datasets used for benchmarking vision techniques contain unintended bias, which often gets amplified by deep learning-based models \cite{hendricks2018women, Stock_ECCV2018, Wang_2019_ICCV}. Therefore, it is imperative for the training data to be analyzed in order to uncover potential biases or unexpected distributional artifacts. 

Semantic context among objects often serves as a valuable cue for scene understanding. Specific relations among the objects help our mind to visualize a scene \cite{biederman1982scene}. Similarly, deep convolutional networks implicitly learn representations encoding contextual relations between objects that help in solving the underlying tasks \cite{tang2015improving,barnea2019exploring}. However, unintended, and biased contextual relationships implicitly learned by deep models can also lead to failure while detecting objects outside their obvious context \cite{rosenfeld2018elephant}. 

Contextual bias occurs when certain co-occurring, but spuriously present/learnt relationships, influence the decision-making of a DNN model. Wang et al. \cite{revise} have identified that existing datasets suffer from contextually biased relations; for example, in MS-COCO, women occur in more indoor scenes and with kitchen objects, whereas men co-occur more with sports objects, and in outdoor locations. Contextual bias like above may impact a model's performance in detecting an object with a protected attribute out of its context, e.g., higher error in predicting women with skis in an outdoor setting. 
While collecting new datasets with reduced bias can be a solution, a more widely applicable, and effective solution could be algorithmic curation of existing datasets that lead to trained models which generalize well despite objects appearing in unseen contexts. 

Recent works \cite{context_cvpr2020} have proposed algorithmic changes to handle co-occurring bias in the dataset, which restricts its scope to be re-integrated in every application. An alternate approach for bias mitigation is to \emph{repair} the training data before feeding it to a machine learning model \cite{li2019repair}. These approaches only update the training set and can readily be incorporated into any machine learning model: irrespective of its task or architecture, and by simply retraining the model on the newly curated training set. 

Dataset repair approaches are often criticized for possible data reduction due to re-sampling \cite{context_cvpr2020}. However, in most dataset curation tasks, the bottleneck is often in the data annotation and not the data capturing stage. It may be prudent to capture a large dataset in these scenarios and then resample a fair subset from it for annotation, using unsupervised or active learning (AL) approaches. Even in the scenarios where data capturing is costly, dropping samples to repair the dataset may not always significantly deteriorate the model accuracy, as has been observed by recent AL techniques \cite{sinha2019variational,cdal,coreset}. 
Inspired by these advances in AL, we argue that it is possible to devise a resampling strategy that can maintain the model accuracy while substantially reducing different sources of bias in a dataset. 

Our resampling approach is inspired by the inequality measures like the Generalized Entropy Index (GEI) used to capture income inequality in populations \cite{Cowell_2003}. Specifically, we use a special case of GEI, the \emph{coefficient of variation} ($ c_v $) statistic, as a measure of inequality of representation of co-occurring classes in the dataset. By following a sampling strategy that minimizes this statistic in subsequent active learning iterations, we demonstrate a reduction in co-occurrence bias while maintaining the prediction accuracy. Furthermore, we show additional results on different tasks where the proposed sampling strategy can reduce prediction bias. We summarize our contributions below: 
\begin{enumerate}[leftmargin=*]
	\item We introduce a simple but effective data repair algorithm to curate a fair dataset, free from contextual co-occurring bias, using \emph{coefficient of variation ($c_v$)}. Our curation algorithm is shown to be effective in both supervised and unsupervised settings.
	\item In a supervised setting, we show that our fair selection algorithm achieves $c_v$ value of almost 0, thus reducing the representational bias and improving model performance over the baselines techniques \cite{Wang_2019_ICCV, li2019repair, focal_loss, classbalancingloss, context_cvpr2020}.
	\item In an unsupervised setting, we experiment in the active learning configuration and show improvements over the recent state of the art AL techniques \cite{cdal, coreset, iclr2021, luo2013latent} in selecting fair subsets and learning fair models. 
	\item We validate the generalizability of models trained on fairly curated data through a series of different experiments and cross-data evaluation.
\end{enumerate}

\section{Related Work}

\myfirstpara{Identifying Bias in Datasets}
Bias in the datasets is being studied for a while, especially in the field of NLP~\cite{sun2019mitigating, mehrabi2020man, liang2020towards}. Torralba and Efros \cite{2011_efors} pointed out similar risks in visual datasets. As a solution, Datasheets for Datasets \cite{gebru2018datasheets} was proposed, which encouraged the dataset creators to follow a certain protocol while collecting data. 
In recent work~\cite{revise}, the authors propose a tool that investigates bias in a visual dataset based on the object, gender and geographic locations. Wilson et al. \cite{wilson2019predictive} have also identified the demographic stereotype learned by object detectors, showing higher error rate for detecting pedestrians with darker skin tones. As reported in \cite{yang2020towards}, ImageNet, one of the most popular and largest vision datasets also suffers from bias, where the authors examined the ImageNnet person subtree and its demographic bias. 
While some forms of bias can be discovered by analysing raw datasets, it is more difficult to identify bias inherited by deep learning models during training, e.g., representational bias. 
In this work we focus on \emph{contextual bias}, which we define as representational bias arising from a skewed distribution of co-occurring objects. Recent works \cite{context_cvpr2020} have also identified co-occurring bias between a \emph{pair} of classes, and using class activation maps to improve the model's performance in identifying co-occurring object exclusively, i.e., out of its typical context. In this paper, we propose a data sampling technique that is effective in reducing bias for \emph{multiple co-occurring objects}. 

\mypara{Mitigating Bias}
Previously, linear models have been proposed to mitigate the bias in several ML algorithms~\cite{zafar2017fairness, celis2019classification, dwork2012fairness}. More recently, there is a shift towards reducing model bias in an end-to-end manner~\cite{li2018resound,li2019repair,Wang_2019_ICCV,yang2020towards, bahng2020learning,menlikeshopping,wang2020towards, tang2020unbiased,tartaglione2021end,gong2021mitigating,xu2020investigating}. Among these, approaches relying on resampling of data to reduce bias remain scarce, with works often criticizing them with claims that balancing data results in bias amplification \cite{Wang_2019_ICCV} . However, REPAIR \cite{li2019repair} has introduced a sampling technique based on the weight assigned to each sample to reduce representational bias and also showed increase in the model performance. We argue that a well-designed curation technique can be an effective solution for the problem of dataset bias.

\mypara{Active Learning}
AL techniques help select the most uncertain samples which can be annotated to reduce the annotation cost and achieve the best performance with limited data. Various AL scenarios like pool-based \cite{wei2015submodularity, huang2010active, sinha2019variational,yoo2019learning}. query-based \cite{freund1997selective, wang2015querying} selection have been proposed.
More recent works \cite{cdal,sinha2019variational,yoo2019learning} have also included diversity and representativeness as a measure to select informative samples. There are few works that have proposed AL techniques with fairness as one of the objectives. Very recently, \cite{iclr2021} incorporated fairness in the acquisition functions using the existing AL technique \cite{bald} with the goal of training fair models. FAL \cite{fair_active_learning} queries data points in each round of selection expecting a fair model, which creates significant overhead.
On the other hand, our proposed technique curates unbiased samples a-priori and thus trained model is fair and better in terms of underlying task performance.

\section{Methodology}

Inspired by minimizing the inequality for contextual co-occurrence bias, we use the objective function as the \emph{coefficient of variation}, which is a special case of Generalized Entropy Index \cite{DeMaio2007}. We now present the notation and the formulation, followed by our proposed algorithms for selecting fair samples.

\subsection{Problem Formulation}
\label{sec:formulation}

Let $ \mathcal{D} = \{ I_1,I_2,\ldots,I_N \} $ be a dataset of $ N $ images with $ \{ c_1, c_2, \ldots, c_K \} $ as the set of $ K $ object classes present. We define $ \mathbf{C} $ as the binary composition matrix of size $ N \times K $. Each element $ (i,j) $ of the composition matrix indicates the presence (1) or absence (0) of the $ j^{th} $ class in the $ i^{th} $ image. For a certain protected class $c_{\pi}$, we define the constituent  $N_{\pi}\times K-1 $ sub-matrix $\mathbf{C}_\pi$ \footnote{$ \mathbf{C}_\pi $ contains all the rows of $ \mathbf{C} $ for which the entry in the column $ j_{c_\pi} $ corresponding to class $ c_\pi $ is 1, and all columns but $ j_{c_\pi} $. } that represents all $ N_\pi\le N $ images containing the class $ c_\pi $ and the corresponding $ N_\pi $ element subset of $ \mathcal{D} $ as $ \mathcal{D}_\pi $. 

We are interested in selecting a subset $ \mathcal{S} \subseteq \mathcal{D}_\pi $ where all co-occurring classes of $ c_\pi $ are well represented. So we define the binary selection vector $ \mathbf{s} \in \{0,1\}^{N_\pi} $ that indicates the set of images selected from $ \mathcal{D}_\pi $. We obtain the number of images per co-occurring class in $ \mathbf{n}_\pi \in \mathbb{N}^{1\times K}$ as:
\begin{align}
	\mathbf{n}_\pi = \mathbf{s}^\top\mathbf{C}_\pi 
\end{align}
Defining the number of images as a measure of richness of representation, we can compute the Generalized Entropy Index as a measure of inequality among these co-occurring classes:  
\begin{align}
	\text{GEI}(\alpha) = 
	\begin{cases}
	\frac{1}{(K-1)\alpha(\alpha-1)}\sum_{i=1}^{K}\left[\left(\frac{\mathbf{n}_\pi(i)}{\mu_{\tiny \mathbf{n}_\pi}} \right)^\alpha -1 \right], &~~\alpha \neq 0,1\\
	\frac{1}{K-1}\sum_{i=1}^{K-1}\frac{\mathbf{n}_\pi(i)}{\mu_{\tiny \mathbf{n}_\pi}} \ln \frac{\mathbf{n}_\pi(i)}{\mu_{\tiny \mathbf{n}_\pi}}, &~~\alpha = 1\\
	-\frac{1}{K-1}\sum_{i=1}^{K-1}\ln \frac{\mathbf{n}_\pi(i)}{\mu_{\tiny \mathbf{n}_\pi}}, &~~\alpha = 0
	\end{cases}
\end{align}
where $ \mu_{\tiny \mathbf{n}_\pi} = \mathbf{n}_\pi^\top \mathbf{1}/(K-1)$ is the mean number of images per class, and $ \mathbf{n}_\pi(i) $ is $ i^{th} $ element of the vector $ \mathbf{n}_\pi $. $\textbf{1}$ is a $K$ dimensional vector of ones. For $ \alpha=2 $, the GEI becomes the square of the coefficient of variation ($ c_{v} = \frac{\sigma_{\tiny \mathbf{n}_\pi}}{\mu_{\tiny \mathbf{n}_\pi}} $), where $ \sigma_{\tiny \mathbf{n}_\pi} $ is the standard deviation of the elements of $ \mathbf{n}_\pi $. Therefore, we formulate the following optimization problem to sample a subset of size $ \mathcal{B} = |\mathcal{S}| $.
\begin{align}
\label{eq:opt_form}
	\min_{\mathbf{s}\in\{0,1\}^{N_\pi}}  & \quad \frac{(K-1)\mathbf{s}^\top \overline{\mathbf{C}}_\pi\mathbf{s}}{(\mathbf{s}^\top\mathbf{C}_\pi\textbf{1})^2}\\ 
	\nonumber \text{s.t.}  & \quad \mathbf{s}^\top\textbf{1} = \mathcal{B}
\end{align} 
where matrix $ \overline{\mathbf{C}}_\pi$ is given by $ \mathbf{C}_\pi\left(\mathbf{I}-\frac{1}{\!K\!-\!1\!}\textbf{1}\textbf{1}^\top\right)\left(\mathbf{I}-\frac{1}{K-1}\textbf{1}\textbf{1}^\top\right)^\top\mathbf{C}_\pi $. Here $ \mathbf{I} $ is a $ \!K\!-\!1\! \times \!K\!-\!1\! $ identity matrix, and $ \textbf{1} $ is a $ \!K\!-\!1\! $ dimensional vector of ones. The numerator of the objective function contains the expression for the variance of $ \mathbf{n}_\pi $ and the denominator is the squared mean of $ \mathbf{n}_{\pi} $. The variable $\mathbf{s}$ here is the selection vector, which is a binary vector with ones denoting the indices of images selected. The solution to (\ref{eq:opt_form}) provides the optimal selection that yields the smallest $ c_{v} $ in the selected subset.
We emphasize that our focus is to show that a sound resampling strategy is effective in mitigating model bias. We note that it is possible to use a more sophisticated algorithm to solve (\ref{eq:opt_form}) optimally. However, we show empirically that even a simple, greedy approach for sample selection produces desirable results. Next, we describe our approach to solve (\ref{eq:opt_form}) for two scenarios: first, the fully supervised setting, and second, in the active learning setting. The former permits us to eliminate co-occurrence bias from existing datasets, while the latter allows us to curate fair datasets. 


\subsection{Supervised Fair Selection}
\cref{alg:fair_select} describes the selection of a subset of images $\mathcal{S}$ given $\mathbf{C}_\pi$ corresponding to the protected class $c_\pi$, while minimizing $c_v$.
 
\begin{algorithm}
    \caption{Fair Selection}
    \label{alg:fair_select}
	\textbf{Input:} $ \!\mathbf{C}_\pi\!  \in\! \{0,1\}^{N_\pi\times K-1}$, Budget $b,\mathcal{S}\! =\! \{\}$\\
	\textbf{Output:} $\mathcal{S}$ 
    \begin{algorithmic}[1]
	\State $\mathcal{S} \sim \cup ~\mathbf{C}_\pi$
    \State $\mathbf{C}_\pi' = \mathbf{C}_\pi \setminus \mathcal{S}$
    \State \textbf{repeat}
    \State $\>$ $\>$ $min = \infty ; k = \infty$
    \State $\>$ $\>$ \textbf{for} $y$ in $\mathbf{C}_\pi'$
    \State $\>$ $\>$ $\>$ $\>$ $ r_{y} = c_v(\mathcal{S} \cup \{y\})$
    \State $\>$ $\>$ $\>$ $\>$ $\textbf{if}$ $ r_{y} < min$
    \State $\>$ $\>$ $\>$ $\>$ $\>$ $\>$ $ min = r_{y} ; k = y$
    \State $\>$ $\>$ $\mathcal{S} = \mathcal{S} \cup \{k\}$
    \State $\>$ $\>$ $\mathbf{C}_\pi' = \mathbf{C}_\pi' \setminus \{k\}$
    \State \textbf{until} $|\mathcal{S}| < b$
    \State \textbf{return} $\mathcal{S}$
    \end{algorithmic}
\end{algorithm}

\subsection{Active Learning for Fair Training (ALOFT)}
In this section we describe our data curation approach, Active Learning for Fair Training (ALOFT). We begin with an initial unlabeled pool of data $\mathcal{D}_{\pi}^{U}$ for a protected class $\mathbf{c}_{\pi}$, and a base model $\mathcal{M}$. As with typical AL methods, ALOFT also uses a sampling budget $b\!=\!\vert\mathcal{B}\vert$ in each cycle. The ALOFT acquisition function selects a subset $\mathcal{B} $ of samples from $\mathcal{D}_{\pi}^{U}$ for which the annotations are requested from a human oracle. The samples in $\mathcal{B}$ are removed from the $\mathcal{D}_{\pi}^{U}$ and are added to the labeled set $\mathcal{L}$ along with the corresponding annotations, which are then used to retrain the model $\mathcal{M}$. As opposed to the traditional usage of a measure of uncertainty or diversity or both, ALOFT minimizes the coefficient of variation $c_v$ across co-occurring classes for selecting contextually class balanced samples. Algo. [\ref{alg:aloft}] lists the different steps in a single AL cycle of ALOFT.  


\begin{algorithm}
    \caption{Sampling strategy for ALOFT}
    \label{alg:aloft}
    \textbf{Input:} $\!\mathbf{C}_\pi\!  \in\! \{0,1\}^{N_\pi\times K-1}, b, \mathcal{D}_\pi^U$ \quad\quad\quad\quad\quad\quad\quad\\
    \textbf{Output:} $\mathcal{L}$, $\mathcal{D}_{\pi}^{U\setminus\mathcal{B}}$\quad \quad\quad\quad\quad\quad\quad\quad\quad\quad\quad\quad\quad\quad\quad\quad\quad\quad\quad\quad\quad
    \begin{algorithmic}[1]
        \State Initialize: $\mathcal{B}= \mathcal{L}=\{\phi\}$
        \State $S_{x}$ = Fair Selection($\mathbf{C}_\pi$, $\mathcal{B}$,$\mathcal{S}$)  \quad\quad\quad\quad\quad\quad Algo[\ref{alg:fair_select}]
        \State $S_{y}$ = $\emph{ORACLE}(S_{x})$
        \State $\mathcal{L}$ = $\mathcal{L}$ $\cup$ $S_{y}$
        \State $\mathcal{D}_{\pi}^{U\setminus\mathcal{B}}$ = $\mathcal{D}_\pi^U \setminus S_{x}$
        \State \textbf{return} $\mathcal{L}$, $\mathcal{D}_{\pi}^{U\setminus\mathcal{B}}$
    \end{algorithmic}
    
\end{algorithm}
To use the fair selection Algo. [\ref{alg:fair_select}] in an active learning setting, we need the binary composition matrix $\mathbf{C}_\pi$. In the absence of ground truth labels while curating a dataset, we construct $\mathbf{C}_\pi$ using the \emph{pseudo-labels} obtained from our base model $\mathcal{M}$. By using pseudo-labels, which in turn rely on the large receptive fields typical in most CNN-based object detectors, we are able to exploit the model's predictive uncertainty as well as implicitly capture the contextual semantic relations between the objects and their representations in the composition matrix $\mathbf{C}_\pi$. 


For a given image $\mathbf{I}$, a typical CNN-based object detector model generates $n_r$ region proposals and a corresponding $|\mathcal{C}|+1$ dimensional\footnote{for $|\mathcal{C}|$ classes and background} softmax probability vector, where $\mathcal{C}$ is the set of classes of interest\footnote{In all our experiments, $|\mathcal{C}|=K-1$, where $c_\pi$ is left out.}. Let $\mathbf{I}_r$ be the $r^{th}$ region in $\mathbf{I}$ and $\mathbf{p}_r$ be the corresponding object class probability vector, which is stacked together to form the probability matrix $\mathbf{P}_\mathbf{I}\in [0,1]^{n_r\times |\mathcal{C}|+1}$. We say that an object class is present in $\mathbf{I}$ if its probability is maximal for \emph{any} of the $n_r$ regions. In other words, we collect the set of classes $\mathcal{C}_\mathbf{I}$ in image $\mathbf{I}$ as 
\begin{align}
    \mathcal{C}_\mathbf{I} = \bigcup_{i\in[n_r]} \arg\max \mathbf{P}_\mathbf{I}[i,\cdot]
\end{align}
where, $[n_r]=\{1,2,\ldots,n_r\}$ and $\mathbf{P}_\mathbf{I}[i,\cdot]$ represents the $i^{th}$ row of $\mathbf{P}_\mathbf{I}$. Using the corresponding sets $\mathcal{C}_\mathbf{I}$ for all images $\mathbf{I}_j, ~j\!\in\![N_\pi]$, we construct $\mathbf{C}_\pi$ for a given unlabeled pool $\mathcal{D}_\pi^U$ in each AL cycle (Algo. [\ref{alg:aloft}]), where the next subset for annotation is selected using (Algo. [\ref{alg:fair_select}]) for labeling. Following this strategy, our labeled set $\mathcal{L}$ is maintained to be contextually balanced, leading to a fair curated dataset. 

\section{Experimental Setup}

\begin{table*}[t]
	\centering
	\footnotesize
	\caption{In this table we report the $c_v$ scores of each selection by different techniques on different budgets. We choose `Cup' as the protected class, and each value in the table is the percentage of instances selected of a particular co-occurring class. Last row, reports the total instances of each object in 100\% data. Closer the $c_v$ to 0, more contextually balanced is the curated data.}
	\vspace{0.1cm}
	\begin{tabular}{|c|l|cccccccccc|c|}\hline
		Data(\%)& \makecell[c]{Method} & Person & Din-Table & Bottle &Chair &Bowl & Knife & Fork & Spoon & Wine Glass & Sink & $c_v(\downarrow)$\\\hline
		\multirow{5}{*}{10} & Random & 0.57 & 0.56 & 0.36 & 0.35 & 0.32 & 0.25 & 0.22 & 0.21 & 0.12 & 0.13 & 0.486\\
		& Ranking & 0.67 & 0.59 & 0.54 & 0.44 & 0.54 & 0.45 & 0.38 & 0.4 & 0.3 & 0.2 & 0.26\\
		& Per-class rank & 0.73 & 0.42 & 0.14 & 0.15 & 0.09 & 0.07 & 0.07 & 0.06 & 0.03 & 0.0 & 1.18\\
		& Threshold & 0.64 & 0.64 & 0.44 & 0.36 & 0.47 & 0.34 & 0.31 & 0.31 & 0.21 & 0.11 & 0.422\\
		& \textbf{Ours} & 0.36 & 0.36 & 0.36 & 0.36 & 0.36 & 0.36 & 0.36 & 0.36 & 0.36 & 0.36 & \textbf{0.0014}\\\hline
		\multirow{5}{*}{20}  & Random & 0.55 & 0.54 & 0.35 & 0.33 & 0.32 & 0.25 & 0.22 & 0.22 & 0.12 & 0.12 & 0.467 \\
		& Ranking & 0.64 & 0.61 & 0.49 & 0.42 & 0.51 & 0.41 & 0.34 & 0.41 & 0.28 & 0.17 & 0.31 \\
		& Per-class rank & 0.67 & 0.51 & 0.21 & 0.32 & 0.21 & 0.1 & 0.09 & 0.11 & 0.04 & 0.0 & 0.91 \\
		& Threshold & 0.64 & 0.64 & 0.43 & 0.38 & 0.46 & 0.35 & 0.3 & 0.33 & 0.24 & 0.13 & 0.401\\
		& \textbf{Ours} & 0.32 & 0.32 & 0.32 & 0.32 & 0.32 & 0.32 & 0.32 & 0.32 & 0.32 & 0.32 & \textbf{0.0008}\\\hline
		\multirow{5}{*}{30}  & Random & 0.57 & 0.54 & 0.34 & 0.34 & 0.32 & 0.24 & 0.22 & 0.22 & 0.13 & 0.12 & 0.469 \\
		& Ranking & 0.63 & 0.62 & 0.46 & 0.39 & 0.51 & 0.37 & 0.31 & 0.36 & 0.35 & 0.15 & 0.35 \\
		& Per-class rank & 0.64 & 0.54 & 0.28 & 0.35 & 0.23 & 0.11 & 0.11 & 0.1 & 0.06 & 0.01 & 0.83 \\
		& Threshold & 0.62 & 0.69 & 0.43 & 0.38 & 0.46 & 0.34 & 0.29 & 0.33 & 0.21 & 0.13 & 0.398\\
		& \textbf{Ours} & 0.32 & 0.32 & 0.33 & 0.31 & 0.32 & 0.31 & 0.31 & 0.32 & 0.31 & 0.31 & \textbf{0.017}\\\hline
		\multirow{5}{*}{40}  & Random & 0.57 & 0.54 & 0.34 & 0.34 & 0.32 & 0.23 & 0.22 & 0.22 & 0.12 & 0.12 & 0.473 \\
		& Ranking & 0.62 & 0.63 & 0.44 & 0.38 & 0.47 & 0.35 & 0.3 & 0.34 & 0.23 & 0.13 & 0.38 \\
		& Per-class rank & 0.65 & 0.56 & 0.29 & 0.37 & 0.26 & 0.13 & 0.13 & 0.12 & 0.07 & 0.01 & 0.76\\
		& Threshold & 0.62 & 0.63 & 0.43 & 0.38 & 0.46 & 0.34 & 0.29 & 0.33 & 0.22 & 0.13 & 0.394 \\
		& \textbf{Ours} & 0.34 & 0.36 & 0.34 & 0.31 & 0.33 & 0.31 & 0.29 & 0.31 & 0.27 & 0.28 & \textbf{0.08}\\\hline
		\multirow{5}{*}{50}  & Random & 0.57 & 0.55 & 0.34 & 0.34 & 0.32 & 0.23 & 0.22 & 0.22 & 0.13 & 0.13 & 0.47 \\
		& Ranking & 0.62 & 0.63 & 0.43 & 0.37 & 0.45 & 0.33 & 0.28 & 0.31 & 0.21 & 0.12 & 0.41 \\
		& Per-class rank & 0.64 & 0.56 & 0.31 & 0.36 & 0.27 & 0.15 & 0.15 & 0.14 & 0.08 & 0.03 & 0.698\\
		& Threshold & 0.62 & 0.61 & 0.44 & 0.35 & 0.40 & 0.38 & 0.28 & 0.33 & 0.21 & 0.15 & 0.395\\
		& \textbf{Ours} & 0.36 & 0.38 & 0.34 & 0.31 & 0.33 & 0.31 & 0.28 & 0.28 & 0.23 & 0.25 & \textbf{0.14}\\\hline
		100                & - & 0.57 & 0.55 & 0.35 & 0.34 & 0.33 & 0.23 & 0.22 & 0.22 & 0.13 & 0.13 & 0.49\\\hline
	\end{tabular}
	\label{tab:cv_cup_supervised}
\end{table*}

\mypara{Dataset}
We have used two of the largest object detection dataset COCO \cite{coco} and OpenImages \cite{openimages} for validating performance of models trained using our curated datasets for object detection task. We report all our findings on the COCO 2017 validation set, and have used OpenImages, ObjectNet \cite{objectnet} and Dollar Street\footnote{https://www.gapminder.org/dollar-street} for cross dataset experiments. ObjectNet resembles real-life scenarios for 113 object categories with objects appearing in unusual contexts. On the other hand, Dollar Street was created with an idea of \textit{``what if the world lives on the same street''}. It consists of 138 different categories showcasing the socio-economic bias between the daily household objects. For gender bias, we have used COCO and CelebA\cite{celebA}, a celebrity face image dataset with 40 attributes. To obtain gender annotations in COCO, we considered the dataset suggested by \cite{menlikeshopping}, and use ground truth captions given for each image to annotate an instance of a person as a man or a woman.

%

\mypara{Evaluation Metrics}
To measure the fairness of selection, we used the coefficient of variation $c_v$ defined in \cref{sec:formulation}. For a perfectly balanced set $c_v=0$. To measure model bias, we used representational bias \cite{li2019repair}: $\mathcal{B}(\mathcal{D},\phi) = \frac{I(Z,Y)}{H(Y)}$, where $I(Z,Y)$ is mutual information between $Z$ and $Y$ and $H(Y)$ is the entropy of $Y$, used as a normalizing term. The metric takes values in $[0,1]$, and characterizes the reduction in uncertainty for the class label $Y$ in presence of feature $Z$. We also use Bias Amplification \cite{Wang_2019_ICCV}: $\Delta = \lambda_{M} - \lambda_{D}$, defined as the difference between data ($\lambda_{D}$) and the model ($\lambda_{M}$). leakage
To measure model performance, we use mAP and F1 score. Inspired by the traditional fairness metrics, we also use Equalized Odds \cite{hardt2016equality}, which measures the true positive rate across a group or protected attribute. The metric is formally defined as: $p(\hat{y}|y=1,G=0) = p(\hat{y}|y=1,G=1)$. Here $y$ is a binary outcome, $\hat{y}$ is the corresponding binary prediction and $G$ is the group index for the particular sample. In our case for a protected class $c_\pi$, the group $G$ corresponds to the $K-1$ co-occurring classes, and for bias reduction we want the disparity in true positive rate for $K-1$ co-occurring classes to be minimum. Hence, to measure bias in a multi-label scenario, we propose a new score \textbf{Disparity in Equalized Odds}, 
$
	\text{EoD} = \sigma(p(\hat{y} | y=1, G=g)), g \in \{ g_1, g_2, \ldots, g_K \}.
$
Here $\sigma$ is a function denoting the variance of all the elements. Note that, for a perfectly unbiased classifier, the variance in true positive rate among different co-occurring classes, and hence the EoD value is 0. 

\section{Experiments and Results}

We re-emphasize the motivation of our work which is to select a subset of dataset, such that the contextual imbalance corresponding to a protected class, measured using $c_v$, is minimum. This in turn should reduce the bias in the predictions made by the model trained using the curated dataset. Through a series of experiments described in this section, we show improvement in both of the above aspects achieved by our method compared to other competitive approaches.

\subsection{Curating Fair Data in Supervised Settings}
\label{sec:supervised_exp}
For contextual fairness in object detection in supervised setting, we created a subset of images having `Cup' as a protected class and its $10$ co-occurring classes like `Person', `Dining Table', `Bottle', `Chair', `Bowl', `Knife', `Fork', `Spoon', `Wine Glass' and `Sink' in COCO dataset. The processed dataset consists of 8459 and 360 images in the training and testing set, respectively, with a $c_v$ of $0.49$ and $0.48$.  Table \ref{tab:cv_cup_supervised} (last row) reports the distribution of co-occurring classes in the training set. 

For comparing the predictive bias in downstream task, we choose the task of object detection. For the comparison, we fine-tune a pre-trained FasterRCNN \cite{frcnn} model using  ResNet50 backbone for $30$ epochs with a batch size of $4$. We used Adam optimizer with a learning rate of $1e^{-4}$. The step size for learning scheduler was set to $5$ with $\gamma=0.5$. In the first set of experiments, we would like to validate our fair selection algorithm and its impact on training the model in a supervised setting. We compare by computing $c_v$ of the selected subsets at different budgets, sampled by different sampling techniques like \textit{Random} selection, \textit{Ranking} (samples of largest weights) based selection, selection based on \textit{per class Rank} (samples of largest weight from each class), and \textit{Threshold} based selection (samples with weight $>0.5$). For each of these competitive approaches we use weights for the each sample learned using the objective function proposed by Repair \cite{li2019repair}. 

\begin{table}[t]
	\footnotesize
	\centering
	\caption{Comparing EoD score and mAP of baseline technique with our proposed selection approach at different budgets in \textbf{supervised setting}.} 
	\vspace{0.1cm}
	\begin{tabular}{|c|l|c|c|c|}\hline
		Data(\%) & \makecell[c]{Sampling Method} & EoD$(\downarrow)$ &mAP$(\uparrow)$  \\\hline
		\multirow{5}{*}{10} & Random  & 0.101 & 26.37 \\
		& Ranking & 0.104 & 28.1 \\
		& Per-class rank & 0.11 & 25.2\\
		& Threshold & 0.118 & 26.27\\
		& Ours   & \textbf{0.086} & \textbf{29.2}\\\hline
		\multirow{5}{*}{20} &Random  & 0.099 & 32.59 \\
		&Ranking & 0.094 & 32.7\\
		&Per-class rank & 0.114 & 28.6\\
		&Threshold & 0.106 & 31.29\\
		&Ours   & \textbf{0.083} & \textbf{34.08}\\\hline
		\multirow{5}{*}{30}&Random  & 0.103 & 35.15 \\
		&Ranking & 0.118 & 36.25\\
		&Per-class rank & 0.118 & 29.8\\
		&Threshold & 0.098 & 34\\
		&Ours   & \textbf{0.079} & \textbf{37.5}\\\hline
		\multirow{5}{*}{40}&Random & 0.102 & 37.6 \\
		&Ranking  & 0.108 & 38.1\\
		&Per-class rank & 0.113 & 30.6\\
		&Threshold  & 0.109 & 34.9\\
		&Ours   & \textbf{0.077} & \textbf{39.22}\\\hline
		\multirow{5}{*}{50}&Random & 0.098 & 38.84 \\
		&Ranking  & 0.11 & 39.5\\
		&Per-class rank  & 0.118 & 35.5\\
		&Threshold & 0.115 & 37.9\\
		&Ours   & \textbf{0.093} & \textbf{40.29}\\\hline
		100 & Original & 0.105 & 47.76\\\hline
	\end{tabular}
	\label{tab:supervised_cup}
\end{table}

\begin{table}[t]
	\footnotesize
	\centering
	\caption{Comparing EoD score and mAP of baseline technique with our proposed selection approach at different budgets in an \textbf{active learning setting}.}
	\vspace{0.1cm}
	\begin{tabular}{|c|l|c|c|c|}\hline
		Data(\%) & \makecell[c]{AL Techniques} & EoD$(\downarrow)$ &mAP$(\uparrow)$  \\\hline
		\multirow{5}{*}{10} & Random  & 0.098 & 21.04 \\
		& Coreset & 0.098 & 21.04 \\
		& MaxEnt & 0.098 & 21.04\\
		& CDAL & 0.098 & 21.04\\
		& ALOFT   & 0.098 & 21.04\\\hline
		\multirow{5}{*}{20} & Random  & 0.110 & 27.83 \\
		& Coreset & 0.110 & 28.7 \\
		& MaxEnt & 0.109 & 29.2\\
		& CDAL & 0.101 & 30.29\\
		& ALOFT   & \textbf{0.083} & \textbf{31.14}\\\hline
		\multirow{5}{*}{30} & Random  & 0.110 & 30.51 \\
		& Coreset & 0.110 & 31.25 \\
		& MaxEnt & 0.116 & 31.69\\
		& CDAL & 0.117 & \textbf{32.75}\\
		& ALOFT   & \textbf{0.084} & 32.45\\\hline
		\multirow{5}{*}{40} & Random  & 0.105 & 31.93 \\
		& Coreset & 0.111 & 32.16 \\
		& MaxEnt & 0.104 & 32.35\\
		& CDAL & 0.105 & 32.12\\
		& ALOFT   & \textbf{0.093} & \textbf{33.95}\\\hline
		\multirow{5}{*}{50} & Random  & 0.105 & 32.84 \\
		& Coreset & 0.105 & 32.72 \\
		& MaxEnt & 0.106 & 32.45\\
		& CDAL & 0.109 & 32.53\\
		& ALOFT   & \textbf{0.098} & \textbf{34.32}\\\hline
		100 & Original & 0.105 & 47.76\\\hline
	\end{tabular}
	\label{tab:ALOFT_cup}
\end{table}


\begin{figure}[h]
	\centering
	\includegraphics[width=\linewidth]{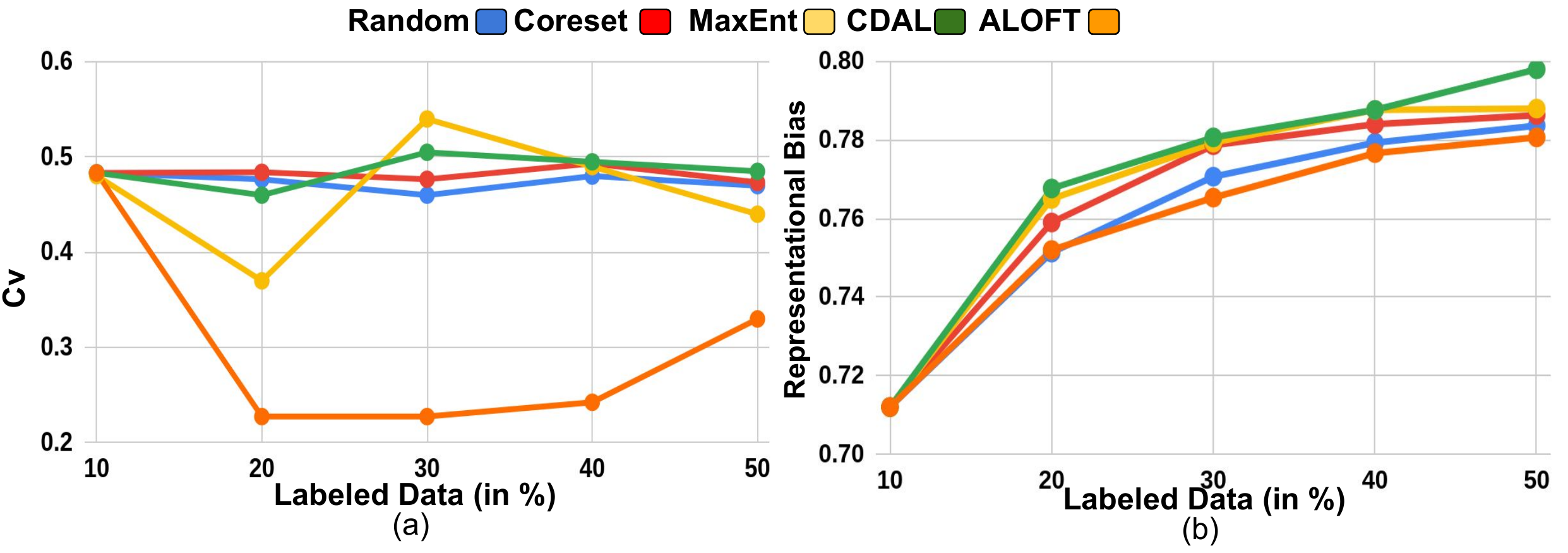}
	\caption{Comparison of ALOFT with other AL baselines. (a)Coefficient of variation ($c_v$) for each selection, lower the value fairer the selection (b)Representational bias of the model. Reported results are avarage of three independent runs.}
	\label{fig:AL_exp}
\end{figure}

In Table \ref{tab:cv_cup_supervised} we summarize the results; we can see that our proposed algorithm selects samples balanced across the different co-occurring classes with a very low $c_v$ value of only $0.0014$ at $10\%$ data as compared to $0.48$ and $0.26$ for `Random' and `Ranking' based selection. Also, note that our selection is highly balanced at a lower budget, but as the budget increases, the $c_v$ also increases due to the limited samples of minor classes. For example, `Sink' has only $0.13\%$  of instances compared to $0.57\%$ of `Person' in $100\%$ training data. Hence, our technique tries to best balance the selection by picking up almost all of it at 50\% budget.

To evaluate the model's improvement in the predictive bias trained using the curated data, we report the EoD score of detecting `Cup' when present along with its various co-occurring classes in \cref{tab:supervised_cup}. We also report mAP values for the trained models in the table. 

\subsection{Curating Data in AL Setting with ALOFT}
\label{sec:aloft_exp}
For the case of unlabeled data, we perform our experiment in an Active Learning setting, considering `Cup and its $10$ contextually co-occurring classes. We begin our experiment with an initially labeled pool of 10\% randomly selected data from the available un-labeled pool and iteratively add samples with a budget of 10\% in each AL cycle. 

%

\mypara{Baselines}
We have compared our ALOFT with following state of the art active learning strategies:
\begin{enumerate}[noitemsep]
	\item \textit{Random Sampling}, for each active learning budget, samples are selected randomly with a uniform probability from the un-labeled pool of data. 
	\item \textit{Coreset} \cite{coreset}, theoretically proven subset selection approach using the feature space of a DNN. We have used their k-center greedy algorithm for comparison. 
	\item \textit{Max Entropy} \cite{luo2013latent}, focuses on selecting the samples for which the model is most uncertain about, by maximizing the entropy of the predicted features.
	\item \textit{CDAL} \cite{cdal}, exploits the contextual confusion among the classes to select contextually diverse samples. 
\end{enumerate}
In \cref{fig:AL_exp}(a) we see that $c_v$ value for ALOFT is fairly low compared to other baselines. We note that although the current AL techniques select diverse and representative samples using contextual information like in \cite{cdal}, but lacks in selecting fair subset. With 20\% data, ALOFT has a $c_v$ value of $0.22$ compared to $0.49$ of Coreset. \cref{fig:AL_exp}(b) reports the representational bias. ALOFT's performance is better than other state-of-the-art. 

In Table \ref{tab:ALOFT_cup} we report the EoD score of detecting `Cup' when appearing with different co-occurring classes.
EoD value closer to $0$ for the model suggests that the learned model has similar accuracy of detecting `Cup' across its context, thus equalizing true positive rate. We also report mAP values computed on the COCO validation by the model on the corresponding curated dataset. 
We see that ALOFT is able to achieve better mAP values than other AL approaches, while keeping the EoD score lower than the rest.

\subsection{Cross Domain Generalization Performance}

Experiments in the previous \cref{sec:supervised_exp,sec:aloft_exp} have shown that a model trained on the data selected using our techniques is balanced (measured using $c_v$), maintains or improve the accuracy (measured by mAP), while reducing the predictive bias (indicated by lower EoD score). In this section we benchmark the generalization of a model trained using data selected by us, but when tested in a cross-domain setting. 

\begin{figure}[t]
	\begin{center}
		\includegraphics[height = 5cm,width =\linewidth]{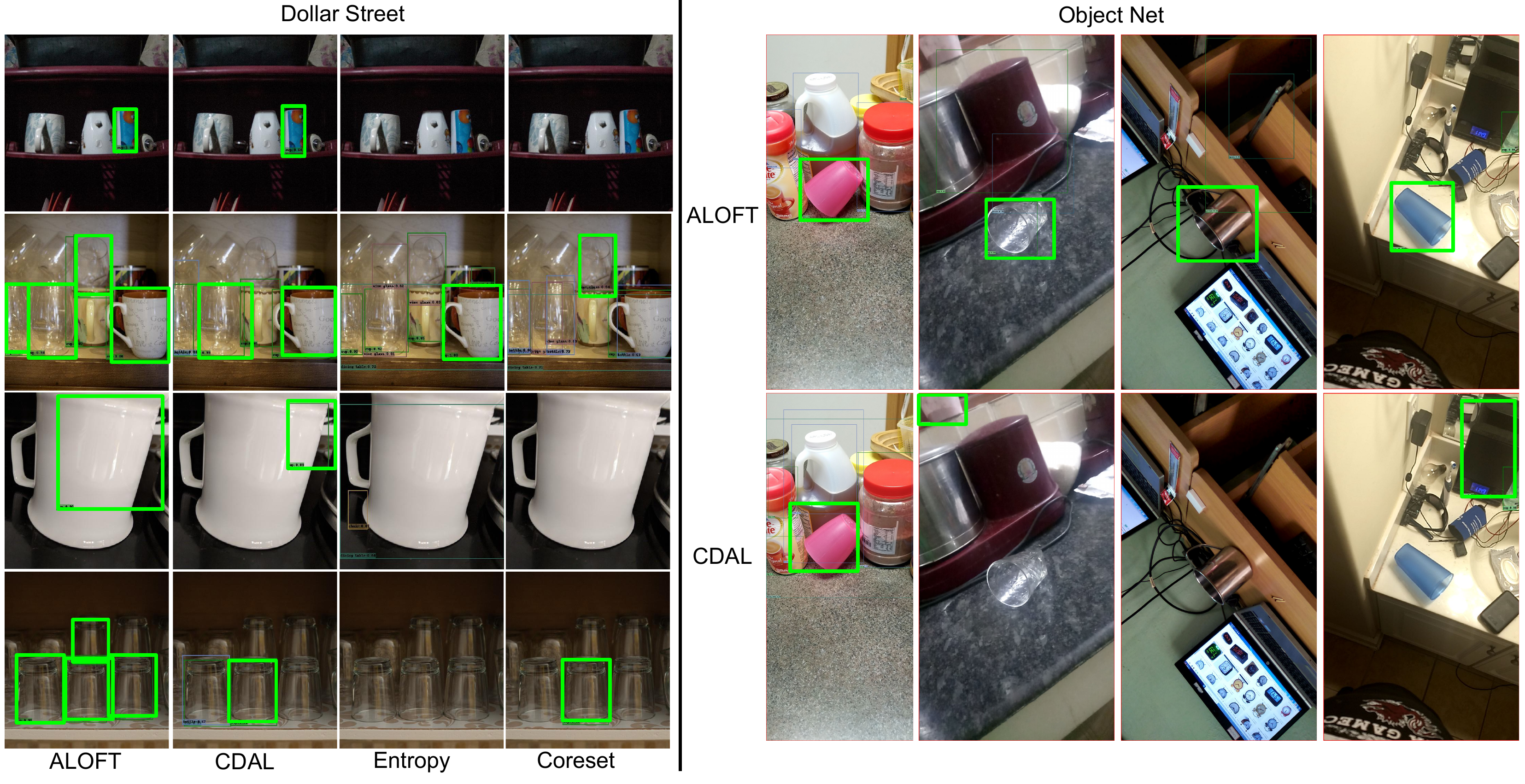}
	\end{center}
	\caption{Qualitative results of detecting `Cup' by different AL techniques on Dollar Street and ObjectNet dataset. Green box denotes predicted bounding box of `Cup' when model trained using only 20\% of training data. (Best visible at 3x zoom) }
	\label{fig:crossdata_qual}
\end{figure}

For this experiment we use the model trained on COCO, but tested on OpenImages \textit{without any finetuning}. Following the experimental setup of previous experiments, we take `Cup' as the protected class, and create a test set from OpenImages having `Cup', and irrespective of its context. Performing this experiment is crucial to check a model's fairness across domains. Each vision dataset has a certain domain representation, which is implicitly learnt by a DNN during training, and impacts its generalizability when tested outside the training domain. From  \cref{tab:cross_dataset} we observe that in both supervised and AL settings, the AP value of `Cup' across all the budgets, improves significantly when compared to other baselines. \cref{fig:crossdata_qual} shows qualitative results of detecting `Cup' by the COCO trained model on ObjectNet and Dollar Street datasets. Since the datasets don't have the ground truth labels for all the objects present in an image, the quantitative experiments could not be performed. 


\begin{table}[t]
	\footnotesize
	\centering
	\caption{Cross Datasset Evaluation. We compare the average precision of detecting Cup on `OpenImages' dataset when model trained on images of Cup from COCO dataset by different methods in supervised(rows 1-3) and active learning setting(rows 4-8).}
	\vspace{0.1cm}
	\begin{tabular}{|c|c|c|c|c|c|}\hline
		\backslashbox{Method}{Data(\%)} & 10 & 20 & 30 & 40 & 50 \\\hline
		Random & 47.7 & 52.6 & 56.97 & 59.4 & 62.2 \\
		REPAIR\cite{li2019repair} & 38.15 & 41.3 & 51.47 & 55.2 & 61.8 \\
		Ours   & \textbf{38.31} & \textbf{55.7} & \textbf{60.9} & \textbf{60.12} & \textbf{64.1}\\\hline
		Random & 41.56 & 48.93 & 52.11 & 55.04 & 57.18\\
		Coreset\cite{coreset}& 41.56 & 27.18 & 31.63 & 35.86 & 48.02 \\
		MaxEnt\cite{luo2013latent} & 41.56 & 46.75 & 37.23 & 51.31 & 57.4\\
		CDAL\cite{cdal} & 41.56 & 45.3 & 37.23 & 49.35 & 45.9\\
		ALOFT  & 41.56 & \textbf{52.13} & \textbf{53.24} & \textbf{56.9} & \textbf{62.5}\\\hline
	\end{tabular}
	
	\label{tab:cross_dataset}
\end{table}

\subsection{Curating Fair Data to Mitigate Gender Bias}

Beyond the contextual bias amongst the objects, we also consider the fairness aspect when a person co-occurs with different objects in an image. We identify the demographic bias in COCO for gender; as sports objects like skateboard, surfboard occurs more frequently with men than women. We also notice that the overall representation of women in COCO dataset is one-third that of men. On the other hand, Women label is still biased favorably in categories like `handbag', `fork', `teddy-bear'. This unequal representation of men and women with different objects leads to intrinsic bias for a model when tested in a real-world setting. 

Wang et al. \cite{Wang_2019_ICCV} also reported this imbalance in COCO and claims that balancing data within a demography would amplify the bias and thus proposed an adversarial de-biasing approach to reduce bias amplification ($\Delta$). We disagree with their claim, and instead propose that it is more effective to balance the data across the demography for reducing the bias. We show that our sampling is more effective in mitigating the bias than their adversarial de-biasing approach. Following their experimental setup of using ResNet-50 on COCO across 79 objects for `male' and `female', we curate data with an $\alpha=1$ such that $1/\alpha < \#(m,y)/\#(w,y) < \alpha$, where $\#(m,y)$ and $\#(w,y)$ denote the number of samples containing men and women with label y. Results on other $\alpha$ values are given in the supplementary. In the table below, we report $c_v$ values of `male' and `female' as the protected and 79 categories as the co-occurring classes. The lower values indicate that our curation better balances gender across its context.
\begin{center}
	\begin{tabular}{|c|c|c|}\hline
		Sampling & $c_v$(male)($\downarrow$) & $c_v$(female)($\downarrow$) \\ \hline
		Balanced\cite{Wang_2019_ICCV} &  1.117 & 0.534 \\
		Ours & \textbf{0.231}&\textbf{0.346} \\ \hline
	\end{tabular}
\end{center}

\cref{tab:balanced_data} summarizes the predictive performance in terms of mAP and F1 scores along with bias amplification score as suggested in \cite{Wang_2019_ICCV}. We have experimented using different classification networks to measure the impact of data curation. We conclude that, our data selection process helps in reducing the bias amplification and also improves mAP.

\begin{table}[h]
	\footnotesize
	\centering
	\caption{We showcase each classification network suffers from bias and thus our selection reduces the bias amplification $\Delta$, and increases mAP and F1 score, when compared with selection heuristic of Balanced data proposed by \cite{Wang_2019_ICCV}. Row(5) Balanced(adv) reports result after applying adversarial debiasing of \cite{Wang_2019_ICCV}.}
	\vspace{0.1cm}
	\begin{tabular}{|c|c|c|c|c|}\hline
		Model & Sampling & $\Delta(\downarrow)$ & mAP$(\uparrow)$ & F1$(\uparrow)$ \\\hline
		\multirow{2}{*}{VGG16\cite{vgg}} & Balanced & 9.8 & 45.56 & 43.39 \\
		& Ours & \textbf{3.5} & \textbf{50.05} & \textbf{44.97} \\\hline
		\multirow{3}{*}{ResNet50\cite{resnet50}} & Balanced & 10.37 & 48.23 & \textbf{42.89} \\
		& Balanced(adv) & 2.51 & 43.71 & 38.98 \\
		& Ours & \textbf{2.3} & \textbf{48.9} & 42.6 \\\hline
		\multirow{2}{*}{MobileNet-V2\cite{mobilenet}} & Balanced & 12.1 & 40.54 & 37.142\\
		& Ours & \textbf{6.78} & \textbf{45.4} & \textbf{38.81} \\\hline
		\multirow{2}{*}{GoogleNet\cite{googlenet}} & Balanced & 14.7 & 39.32 & 33.9\\
		& Ours & \textbf{8.3} & \textbf{44.2} & \textbf{37.3} \\\hline
	\end{tabular}
	
	\label{tab:balanced_data}
\end{table}


\subsection{ALOFT for Mitigating Gender Bias}

\begin{figure}[h]
	\centering
	\includegraphics[width=\linewidth]{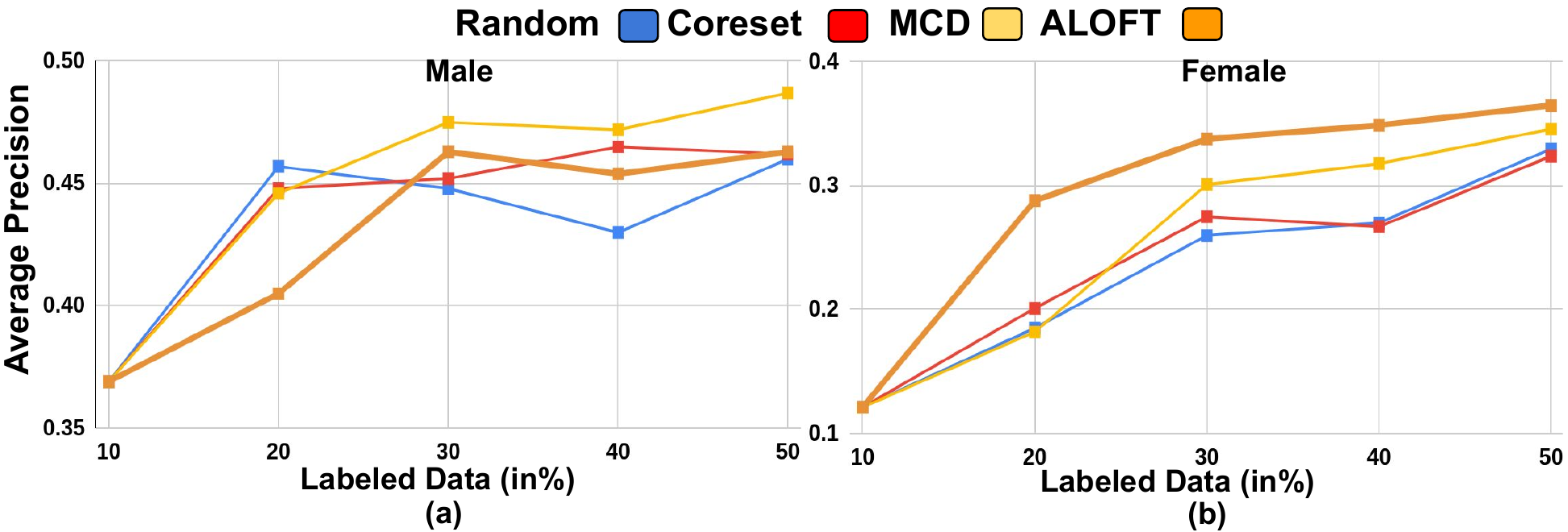}
	\caption{Average precision of detecting (a) Male, (b) Female face in the presence of 39 different attributes, at different budgets. AP of female has significantly improved compromising a bit in Male because of fair sampling.}
	\label{fig:ALOFT_image_classification}
\end{figure}

We perform experiment on CelebA for gender classification in active learning setting. We have compared ALOFT with random, Coreset \cite{coreset}, and MCD \cite{iclr2021}. MCD \cite{iclr2021} proposes a solution to fair training using i.i.d sampling over the existing BALD \cite{bald} based sampling. We consider male/female as the protected class, and other 39 face attributes as co-occurring classes for the selection. We train a ResNet50 \cite{resnet50} based classifier using the selection. In \cref{fig:ALOFT_image_classification}, we report mAP of detecting (a) male and (b) female face in the presence of the other co-occurring 39 attributes.

\subsection{Multi Label Image Classification}

In \cite{context_cvpr2020} authors identified 20 most biased pair of categories in COCO such as \textit{(Skateboard, Person)} where \textit{Skateboard} mostly co-occurs with \textit{Person} than exclusively. Their objective was to improve the performance of the highly biased category when it occurs exclusively. For this experiment, we asked our algorithm to select 31K images out of 80K COCO images such that the number of images of minority class, with and without biased co-occurring class, becomes balanced for all pairs given by \cite{context_cvpr2020}.
Following their experimental setup, we
compare with the one trained on the complete dataset but with a weighted loss which applies 10 times higher weight to the samples of biased category occurring exclusively. The table below shows the comparative performance of the two models in terms of mAP of the minor classes for all 20 pairs. More details and class-wise AP for every biased pair is given in the supplementary material.
%
%
\begin{center}
\begin{tabular}{|c|c|c|}\hline
	Method&Exclusive(mAP)&Co-occur(mAP)\\\hline
	Standard(CE) & 80 & 91\\
	Weighted Loss& 83 & 90.5\\
	Ours & \textbf{84.5} & \textbf{93.5}\\\hline
\end{tabular}
\end{center}
%
%

\section{Conclusion}

We presented a novel, simple yet effective data curation algorithm using the coefficient of variation ($c_v$) that helps to select contextually balanced data for a protected class across its co-occurring biased classes. Our algorithm is effective in selecting contextually fair subsets with a very low $c_v$ value in both supervised and active learning settings. We validate the effectiveness of training a model on contextually balanced data. It helps reduce representational bias and increase the true positive rate for the protected class. We validate that the model's overall performance and generalizability also increases in the cross-dataset setting. For the active learning setting in future we would like to extend our work to explore fairness combined with diversity. We hope our work motivates more effort in addressing different dataset bias and training models on well-curated datasets to make them more reliable and trustworthy.

\mypara{Acknowledgment}
The authors acknowledge the partial support received from the Infosys Center for Artificial Intelligence at IIIT-Delhi. This work has also been partly supported by the funding received from DST through the IMPRINT program (IMP/2019/000250).

{\small
\bibliographystyle{ieee_fullname}
\bibliography{egbib}
}

\end{document}